\documentclass[journal]{IEEEtran}

\usepackage{booktabs}
\usepackage{makecell}
\usepackage{cite}

\usepackage{multirow}
\usepackage{amssymb}

\usepackage[pdftex]{graphicx}

\usepackage{amsmath}

\usepackage{graphicx}
\usepackage{subfigure}
\usepackage{algorithmic}

\usepackage[ruled,vlined]{algorithm2e}

\begin{document}
\title{Efficient Safety Testing of Autonomous Vehicles via Adaptive Search over Crash-Derived Scenarios}

\author{Rui~Zhou}
        % <-this % stops a space
% \thanks{M. Shell was with the Department
% of Electrical and Computer Engineering, Georgia Institute of Technology, Atlanta,
% GA, 30332 USA e-mail: (see http://www.michaelshell.org/contact.html).}% <-this % stops a space
% \thanks{J. Doe and J. Doe are with Anonymous University.}% <-this % stops a space
% \thanks{Manuscript received April 19, 2005; revised August 26, 2015.}}


\markboth{ }%
{Shell \MakeLowercase{\textit{et al.}}: Bare Demo of IEEEtran.cls for IEEE Journals}



\maketitle

\begin{abstract}
Ensuring the safety of autonomous vehicles (AVs) is paramount in their development and deployment. Safety-critical scenarios pose more severe challenges, necessitating efficient testing methods to validate AVs safety. This study focuses on designing an accelerated testing algorithm for AVs in safety-critical scenarios, enabling swift recognition of their driving capabilities. First, typical logical scenarios were extracted from real-world crashes in the China In-depth Mobility Safety Study-Traffic Accident (CIMSS-TA) database, obtaining pre-crash features through reconstruction. Second, Baidu Apollo, an advanced black-box automated driving system (ADS) is integrated to control the behavior of the ego vehicle. Third, we proposed an adaptive large-variable neighborhood-simulated annealing algorithm (ALVNS-SA) to expedite the testing process. Experimental results demonstrate a significant enhancement in testing efficiency when utilizing ALVNS-SA. It achieves an 84.00\% coverage of safety-critical scenarios, with crash scenario coverage of 96.83\% and near-crash scenario coverage of 92.07\%. Compared to genetic algorithm (GA), adaptive large neighborhood-simulated annealing algorithm (ALNS-SA), and random testing, ALVNS-SA exhibits substantially higher coverage in safety-critical scenarios.
\end{abstract}


% \fi
\IEEEpeerreviewmaketitle

\section{Introduction}
\IEEEPARstart{A}{utonomous} driving is widely regarded as a promising technology to enhance road safety. However, safety testing remains one of the major obstacles to the development and deployment of Autonomous Vehicles (AVs). Testing AVs on public roads poses significant risks, and the scenarios encountered are often too simplistic to effectively evaluate AV capabilities \cite{xu2019statistical,huang2024pre,zhou2024evaluating,jin2024analysis,zhou2024would}. As a result, scenario-based virtual testing has emerged as the primary method for AV evaluation due to its efficiency, low cost, and safety.

In scenario-based simulation testing, the core challenge lies in constructing appropriate test scenarios \cite{li2025multidimensional,zhou2022testing}. According to ISO 34501, test scenarios are classified into four levels of abstraction: functional, abstract, logical, and concrete scenarios. Logical scenarios include parameterized descriptions, where some parameters are represented as ranges, while concrete scenarios specify all parameters explicitly to reflect physical properties. Both scenario types have been widely studied. Numerous works have focused on extracting logical scenarios and defining the range of their parameter spaces \cite{lenard2014pedestrian,nitsche2017precrash,sui2021emergency,wang2022autonomous,zhou2023precrash,zhang2025high}. However, there is still no unified approach for generating concrete scenarios from logical ones, despite the fact that concrete scenarios are directly used in testing \cite{huai2023scenorita}. Due to the high dimensionality of scenario parameters, exhaustive exploration of the scenario space is infeasible. Thus, recent research has focused on identifying safety-critical concrete scenarios to accelerate the testing process \cite{abdessalem2018testing,beglerovic2017model,corso2019adaptive,menzel2018scenarios,wang2022gradient,huang2022framework,huang2022functional}.

Most existing accelerated testing methods assume AV control algorithms are white-box systems, with known internal mechanisms that allow for optimized sampling to identify safety-critical scenarios more efficiently \cite{arief2021deep,xu2018accelerated}. However, modern AV control systems are increasingly based on black-box models, typically driven by Artificial Intelligence (AI), which makes previous white-box testing frameworks less applicable \cite{majzik2019towards,zhao2017accelerated}. This shift has led to growing interest in testing methods tailored for black-box AVs, such as unsafe scenario-guided and search-based testing approaches \cite{sun2022scenario,batsch2023taxonomy,gambi2019automatically,zhou2025crash}. These methods aim to rapidly identify high-risk scenarios within a predefined parameter space and rely solely on observed test outcomes, making them suitable for black-box systems. Nonetheless, they often suffer from local optima and limited scenario diversity, which restricts their ability to uncover a broad range of safety-critical scenarios. Therefore, further investigation is required to develop more effective testing strategies for black-box AVs.

Crash scenarios, by their nature, represent high-risk conditions and have been widely utilized in AV testing \cite{cai2020dynamic,zhou2019multi,cai2020fuzzy,cao2019typical,nilsson2018definition,pan2021study}. Extracting representative test scenarios from crash data provides valuable insight into potential safety risks and serves as an effective tool to accelerate AV deployment. Therefore, parameterizing crash scenarios and exploring the corresponding concrete scenario space is of great importance for robust AV testing.

To address the above research gap, this study proposes an \textit{Adaptive Large-Variable Neighborhood Simulated Annealing} (ALVNS-SA) algorithm for accelerated testing of AVs in safety-critical scenarios. First, typical logical scenarios are constructed based on real-world crash data from the \textit{China In-depth Mobility Safety Study–Traffic Accident} (CIMSS-TA) database. The pre-crash characteristics of involved parties are reconstructed to obtain detailed parameter distributions for concrete scenario generation. Next, the black-box AV system Baidu Apollo is used as the ego vehicle controller, and the ALVNS-SA algorithm is applied to accelerate its safety testing. The performance of ALVNS-SA is evaluated in the logical scenario parameter space and compared against the Genetic Algorithm (GA), Adaptive Large Neighborhood Simulated Annealing (ALNS-SA), and random testing approaches.

Experimental results show that ALVNS-SA achieves significantly higher coverage of safety-critical scenarios across all categories compared to baseline methods. The main contributions of this study are as follows:
\begin{enumerate}
    \item A novel approach for constructing logical scenarios from real-world crash data and generating corresponding concrete test scenarios;
    \item The development of ALVNS-SA, an enhanced evolutionary algorithm to accelerate AV safety testing and substantially improve the coverage of safety-critical scenarios.
\end{enumerate}


\section{Related Works}

Accelerated evaluation methods, which aim to improve the efficiency of testing processes, have been widely explored in recent years~\cite{zhu2021hazardous,zhao2025cradle,bian2025search,wei2025generating}. Among them, testing approaches that focus on hazardous or safety-critical scenarios have emerged as a major direction for future research.

Corso et al.~\cite{corso2019adaptive} applied reinforcement learning to identify potential safety-critical scenarios and demonstrated that this data-driven approach can effectively uncover high-risk situations. However, deep reinforcement learning can suffer from challenges such as training instability and slow convergence, which may limit its practical application.

To improve search efficiency, surrogate models have also been introduced. Beglerovic et al.~\cite{beglerovic2017testing} developed a surrogate model to approximate system behavior, which was then optimized using Differential Evolution and Particle Swarm Optimization. Similarly, Mullins~\cite{mullins2018adaptive} proposed an adaptive search algorithm that continuously iterates to minimize sampling effort while identifying the boundaries of hazardous conditions. Wang et al.~\cite{wang2022autonomous} further leveraged surrogate modeling to approximate the safety performance of AVs and applied gradient-based search algorithms to accelerate boundary identification. Nevertheless, the overall effectiveness of these methods is highly dependent on the accuracy of the surrogate model and the efficiency of the optimization algorithm.

Tuncali et al.~\cite{tuncali2016utilizing} introduced S-TaLiRo to automatically generate hazardous test scenarios using simulated annealing. Follow-up studies explored enhancements using gradient descent~\cite{tuncali2017functional} and machine learning techniques~\cite{tuncali2019requirements} to improve convergence speed and robustness. Zhu et al.~\cite{zhu2021hazardous} proposed an optimization-based method to efficiently generate high-risk scenarios. Gambi et al.~\cite{gambi2019automatically} combined procedural content generation with search-based testing to identify dangerous situations. However, both studies primarily focused on individual AV modules—such as adaptive cruise control and lane-keeping systems—which limits the comprehensiveness of the evaluation.

Calò et al.~\cite{calo2020generating} proposed a two-stage method for generating avoidable collision scenarios. This approach first identifies crash-prone scenarios using search-based techniques and then searches for possible avoidance strategies. Batsch et al.~\cite{batsch2019performance} introduced a method based on Gaussian process clustering to identify the performance boundaries of AVs, defined as the transition surface between crash and non-crash outcomes in the parameter space. However, their work did not account for near-crash scenarios, which also present significant safety concerns and warrant further investigation.

A variety of optimization algorithms have also been applied to improve test scenario generation, including Bayesian optimization~\cite{gangopadhyay2019identification}, random forest models~\cite{nabhan2019optimizing}, evolutionary algorithms~\cite{abbas2019safe,abdessalem2018testing,bian2025search}, and reinforcement learning~\cite{ding2020learning,wei2025generating,wei2024risk,luo2025high}.

In summary, while a wide range of optimization techniques have been proposed for the identification and generation of safety-critical scenarios in AV testing, many existing methods face limitations such as redundant exploration, low efficiency, vulnerability to local optima, and instability in training. Additionally, although surrogate models and modular testing can improve computational efficiency, they often reduce realism and overlook system-level behaviors. To address these limitations, this study adopts counterfactual simulation to evaluate the performance of black-box AV systems, enabling a more realistic and comprehensive assessment. Furthermore, we introduce the Adaptive Large-Variable Neighborhood Simulated Annealing (ALVNS-SA) algorithm to efficiently identify high-risk scenarios and enhance testing coverage.
algorithm to optimize the testing efficiency of safety-critical scenarios based on crash logic scenarios. 

\section{Data}

\begin{figure}[htbp]
    \centering
    \includegraphics[width=0.45\textwidth]{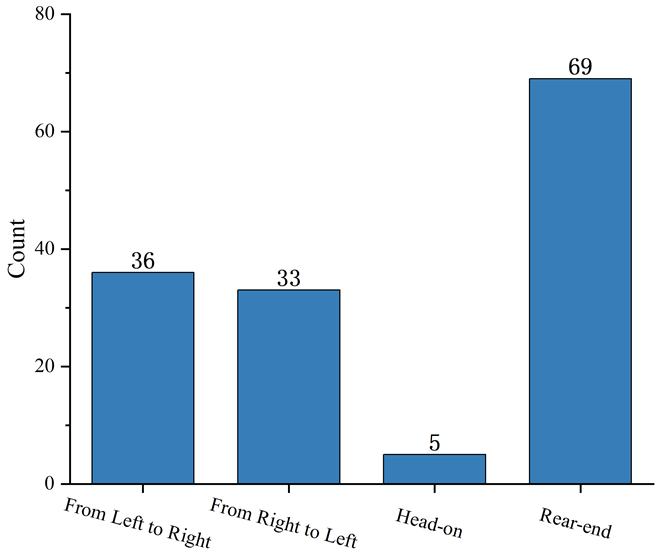}
    \caption{Crash type distribution}
    \label{tab:Crash type distribution}
\end{figure}

This study relies on the CIMSS-TA database, an in-depth crash analysis research project on road safety and AVs safety testing launched by Central South University in 2017. Each case involves at least one vehicle and will be recorded by complete videos through on-road surveillance cameras or onboard video recorders. In order to be consistent with the kinematic model of AVs in the simulation system, only passenger vehicles (sedan, van, SUV, MPV) are considered in this study. Therefore, we filtered out crashes two passenger cars and selected a total of 143 crashes covering the period of 2017 to 2023. Fig.~\ref{tab:Crash type distribution} illustrates the distribution of crash types. This study focuses on investigating rear-end crash, which is the most dominant type of crashes.

\begin{figure}[htbp]
    \centering
    \includegraphics[width=0.45\textwidth]{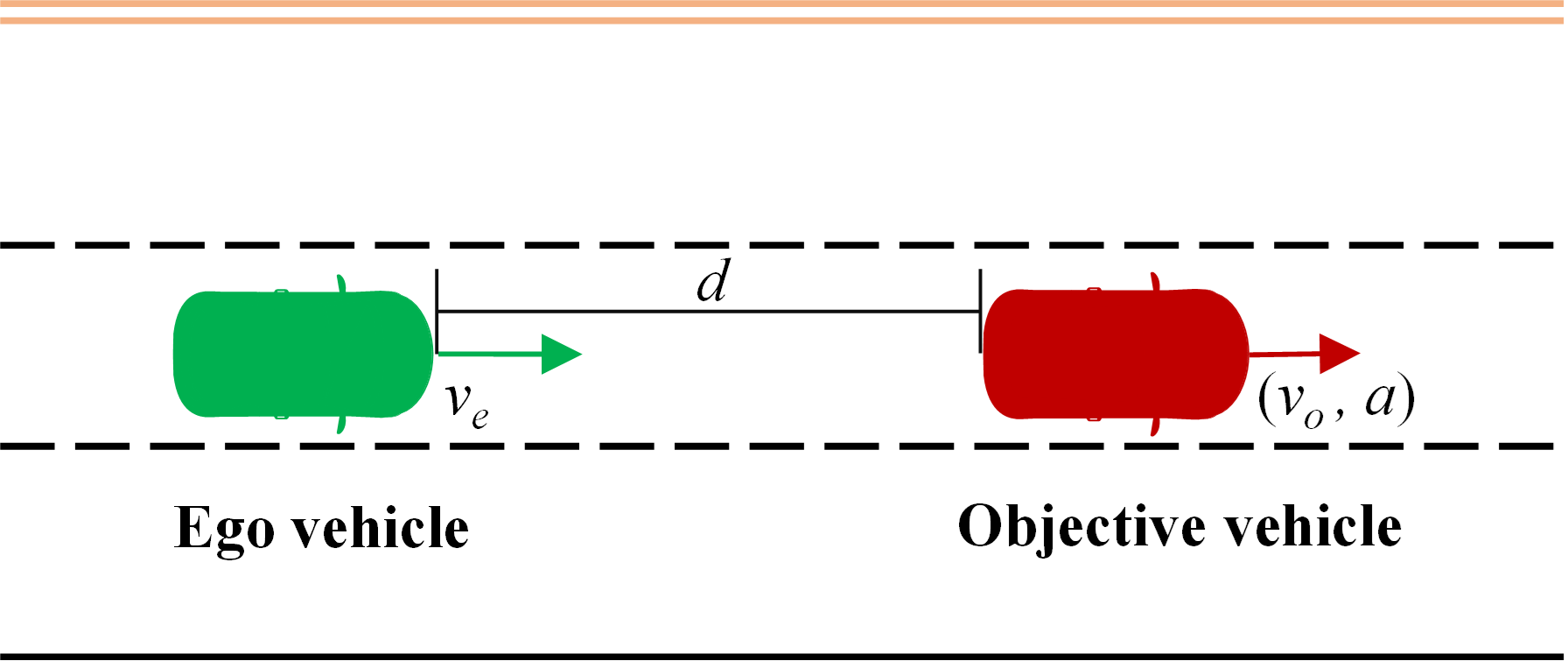}
    \caption{The typical logic scenario}
    \label{fig:The typical logic scenario}
\end{figure}

The relative motion between the ego vehicle and objective vehicle in logical scenarios is designed based on the crash type. However, the basic simulation environmental factors are also indispensable. The logical scenario for this study was constructed by selecting the scenario element with the highest proportion from each category, as shown in Table.~\ref{tab:scenario_elements}. Specifically, the selected logical scenario consisted of clear weather conditions with daytime lighting. It took place on a straight road with six lanes, dry road conditions, no signal control, and no visual obstructions between the ego vehicle and the objective vehicle. as shown in Fig.~\ref{fig:The typical logic scenario}.

\begin{table}[htbp]
\caption{Statistics of scenario elements}
\centering
\scriptsize
\begin{tabular}{llcc}
\toprule
\textbf{Scenario Elements} & \textbf{Category} & \textbf{Count} & \textbf{Proportion} \\
\midrule
\multirow{4}{*}{Road Segment Type}
    & Four-leg intersection & 37 & 25.87\% \\
    & Road entrance and exit & 10 & 6.99\% \\
    & Straightway & 74 & 51.75\% \\
    & Three-leg intersection & 22 & 15.38\% \\
\midrule
\multirow{4}{*}{Number of Lanes}
    & Four-lane road & 67 & 46.85\% \\
    & Single-lane road & 4 & 2.80\% \\
    & Six-lane road or higher & 49 & 34.27\% \\
    & Two-lane road & 23 & 16.08\% \\
\midrule
\multirow{4}{*}{Road Surface Condition}
    & Dry & 98 & 68.53\% \\
    & Freeze & 2 & 1.40\% \\
    & Waterlogged & 18 & 12.59\% \\
    & Wet & 25 & 17.48\% \\
\midrule
\multirow{2}{*}{Traffic Control Method}
    & Signalized & 34 & 23.78\% \\
    & Unsignalized & 109 & 76.22\% \\
\midrule
\multirow{2}{*}{Visual Obstruction}
    & No & 111 & 77.62\% \\
    & Yes & 32 & 22.38\% \\
\midrule
\multirow{3}{*}{Light Condition}
    & Daytime & 103 & 72.03\% \\
    & Night with street lights & 26 & 18.18\% \\
    & Night without street lights & 14 & 9.79\% \\
\midrule
\multirow{3}{*}{Weather}
    & Cloudy & 25 & 17.48\% \\
    & Rainy/foggy/snowy & 24 & 16.78\% \\
    & Clear & 94 & 65.73\% \\
\bottomrule
\end{tabular}
\label{tab:scenario_elements}
\end{table}

\section{Method}

\subsubsection{Crash Reconstruction for Parameter Extraction}

The pre-crash process is simulated and reconstructed using crash reconstruction software, PC-Crash. The final states of the crash-involved parties, including their relative positions and speed parameters, are determined based on the crash records. Based on the video, we utilize the Temporal Difference method and vehicle kinematics to estimate the driving speed of each crash-involved party, which can be recognized by comparing its position between two frames:
\begin{equation}
    v = \Delta l / nt \tag{1}
\end{equation}
\begin{equation}
    t = 1 / fps \tag{2}
\end{equation}
where $\Delta l$ (m) is the distance between two reference points; $n$ is the number of frames required for the vehicle to pass through the two reference points; $t$ (s) is the interval time between each frame; $fps$ is the frame rate. The time-series information such as orientation, yaw rate, position, velocity, and acceleration are obtained through crash reconstruction.

\begin{figure}[htbp]
    \centering
    \subfigure[]{
        \includegraphics[width=0.35\textwidth]{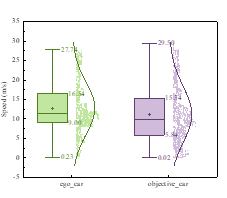}
    }
    \subfigure[]{
        \includegraphics[width=0.35\textwidth]{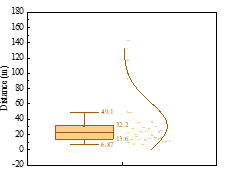}
    }
    \subfigure[]{
        \includegraphics[width=0.35\textwidth]{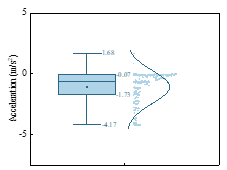}
    }
    \caption{Distribution of parameter: (a) speed; (b) distance; (c) acceleration}
    \label{fig:parameter_space}
\end{figure}

The ego vehicle refers to the AV, which serves as the central element in the testing process and requires thorough assessment and validation for safety and performance. In this study, the rear vehicle is considered the AV. The objective vehicle refers to the primary interacting entity with the ego vehicle during the testing, which specifically denotes one of the parties involved in the crash. The typical logic scenario can be represented by four-dimensional parameters: the speed of the ego vehicle ($v_e$), the speed of the objective vehicle ($v_o$), the distance between the vehicles ($d$), and the acceleration of the objective vehicle ($a$). Thus, a concrete scenario can be denoted as:
\[
    s = (v_e, v_o, d, a)
\]
The parameter space is shown in Fig.~\ref{fig:parameter_space}.

After removing outliers beyond the upper and lower quartiles, the distribution ranges of each parameter are as follows:
\begin{align}
    v_e &\in [9, 16.5] \, \text{m/s} \notag \\
    v_o &\in [5.5, 15.5] \, \text{m/s} \notag \\
    d   &\in [13.5, 32.5] \, \text{m} \notag \\
    a   &\in [-0.05, -1.85] \, \text{m/s}^2 \notag
\end{align}

It is worth noting that the deceleration of the objective vehicle is not constant but is sampled at each time interval following a normal distribution $N(a, \sigma^2)$ to ensure that the trajectory of the objective vehicle is closer to reality. Based on the accuracy of the sensors and the situation of the autonomous driving system, the step sizes were set as $\gamma^{v_e} = 0.5$, $\gamma^{v_o} = 0.5$, $\gamma^d = 1$, and $\gamma^a = 0.2$, respectively. After recombining the parameters, a total of 60,480 concrete scenarios were obtained.

%\IEEEpubidadjcol

\subsection{Safety Assessment of Testing Scenarios}

Generalized Time-To-Collision (GTTC) was used to evaluate the driving risk of AVs. According to the study of Zhang et al. (2023), the simplified GTTC is calculated as follows:

\begin{equation}
    D_{i,j} = \sqrt{(p_i - p_j)^T (p_i - p_j)}
    \label{eq:D_ij}
\end{equation}

\begin{equation}
    D'_{i,j} = \frac{(p_i - p_j)^T (v_i - v_j)}{D_{i,j}}
    \label{eq:D'_ij}
\end{equation}

\begin{equation}
    \text{GTTC} = -\frac{D_{i,j}}{D'_{i,j}}
    \label{eq:GTTC}
\end{equation}

where $p_i, p_j$ are the position vectors of the two closest points on the two vehicles at a given time step, and $v_i, v_j$ are their respective velocity vectors. In Eq.~\eqref{eq:GTTC}, GTTC is calculated only when $D'_{i,j} < 0$. A value of $D'_{i,j} \geq 0$ indicates the crash risk is reducing or constant, as the distance between vehicles is increasing or stable.

A larger GTTC value indicates a safer driving condition. We define $GTTC_{min}$ as the minimum GTTC during the scenario, representing the riskiest moment. Based on $GTTC_{min}$, scenarios are divided into five levels (see Table~\ref{tab:GTTC_levels}). Scenarios with $GTTC_{min} \leq 2.0$ are considered safety-critical and used in the fitness function $f(s)$ for scenario $s$.

\begin{table}[htbp]
\caption{Scenario Classification Based on $GTTC_{min}$}
\label{tab:GTTC_levels}
\centering
\scriptsize  % 缩小字体
\renewcommand{\arraystretch}{1.1}  % 行间距微调
\begin{tabular}{|c|l|p{3.1cm}|}
\hline
\textbf{$GTTC_{min}$} & \textbf{Categories} & \textbf{Description} \\
\hline
$GTTC_{min} = 0$ & Crash scenario & The AV collides with the objective vehicle. Currently, AVs struggle to handle such scenarios. \\
\hline
$0 < GTTC_{min} \leq 0.5$ & Near-crash scenario & The AV comes close to colliding with the objective vehicle but manages to avoid it, representing the safety performance boundary of the AV. \\
\hline
$0.5 < GTTC_{min} \leq 1.0$ & High-risk scenario & There is a severe conflict between the AV and the objective vehicle. It is designed to test the response and decision-making capabilities of AVs under challenging conditions. \\
\hline
$1.0 < GTTC_{min} \leq 2.0$ & Risk scenario & There is a minor conflict between the AV and the objective vehicle. It aims to test the stability of the response and decision-making capabilities of the AV under challenging conditions. \\
\hline
$2.0 < GTTC_{min}$ & Risk-free scenario & This scenario represents a controlled and safe environment where the risk of crashes or adverse events is minimized or eliminated. \\
\hline
\end{tabular}
\end{table}

\subsection{Counterfactual Simulation}

\begin{figure}[htbp]
    \centering
    \includegraphics[width=0.48\textwidth]{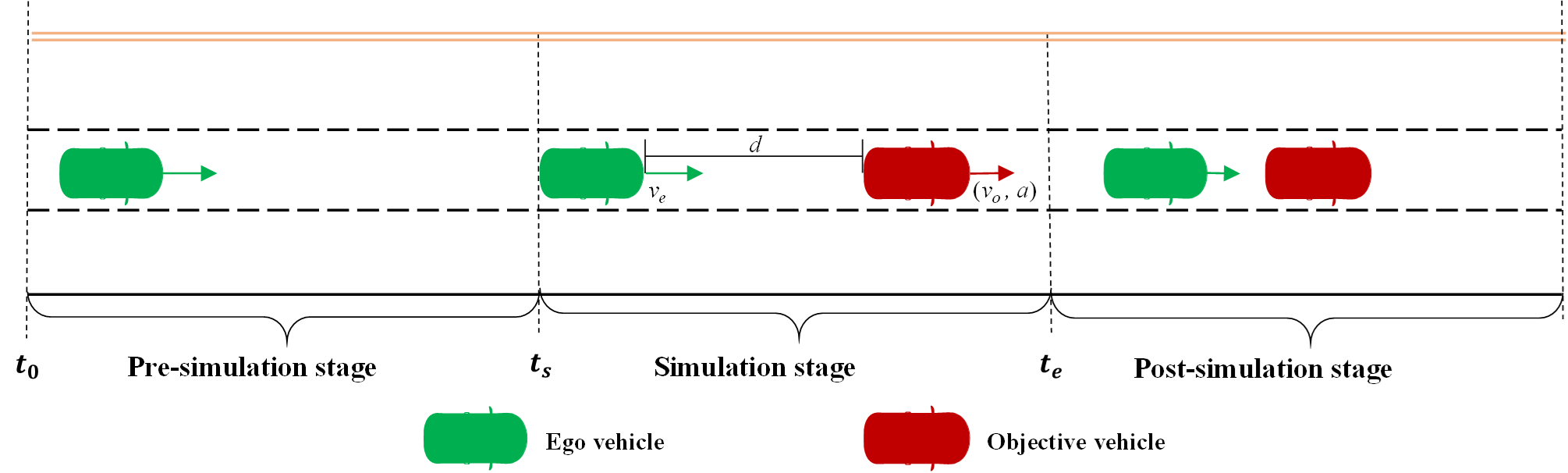}
    \caption{Crash-based counterfactual simulation}
    \label{fig:counterfactual}
\end{figure}

As illustrated in Fig.~\ref{fig:counterfactual}, counterfactual simulation starts by establishing a baseline scenario. Subsequently, simulations are performed to envision potential outcomes under specific counterfactual conditions. For example, in cases involving AVs, simulations can be executed to evaluate the effects of the AV's presence, different decisions, or alternative actions during critical situations. Firstly, the initialization of the simulation is accomplished in SVL Simulator using trajectories of the objective vehicle ($t_0 \sim t_s$). The AV commences recording its motion and sensor data from its initial position of the planned trajectory. Next, the scenario is simulated forward in time ($t_s \sim t_e$): the master vehicle arrives at a preset state, and the objective vehicle follows its predetermined trajectories. Finally, the simulation results are output (from $t_e$ onwards).

\subsection{Accelerated Testing Based on ALVNS-SA}
\subsubsection{Framework of ALVNS-SA}

Counterfactual simulation is used to evaluate the safety of AVs by exploring the consequences of different decision-making scenarios in critical situations. Suppose there is an AV model $A$ that shows and safety requirements $\mathcal{P}$. The external input is a concrete scenario $s \in \mathcal{S}$, where $\mathcal{S}$ is the logic scenario space. Counterfactual simulation means determining whether $\mathcal{P}$ is satisfied for set of crash scenarios $s$ and looks for some concrete scenario set $\Omega_2$ where $A$ violates $\mathcal{P}$:

\begin{equation}
    \textbf{check } \phi(A, s) \models \mathcal{P} \text{ for } s \in \mathcal{S}
\end{equation}
\begin{equation}
    \textbf{find} s \in \Omega_2 \text{ so that } \phi(A, s) \nvDash \mathcal{P}
\end{equation}

The primary objective of this study is to quickly identify $\Omega_2$ and, consequently, expedite the recognition of the capability boundaries of AVs.

Large neighborhood search (LNS) was initially proposed by \cite{shaw1998using} to improve the objective value through the application of destroy-and-repair methods in each iteration, starting from an initial solution. Adaptive large neighborhood search (ALNS) further enhances LNS by incorporating multiple destroy-and-repair methods. Compared to LNS, ALNS allows for more substantial modifications to the current solution by exploring a broader search space. The selection probability of the neighborhood is dynamically controlled based on the historical performance during the search process~\cite{pisinger2010large, ropke2006adaptive, shi2023adaptive}.

Based on the characteristics discussed above, we present a novel approach, namely ALVNS-SA, to expedite scenario-based testing for AVs. ALVNS-SA integrates the destroy-and-repair procedure of ALNS and incorporates the variable neighborhood search (VNS) in the repair process. Additionally, the acceptance criterion for scenarios in ALVNS-SA is governed by the Simulated Annealing (SA) algorithm, a renowned method that emulates the cooling process of a set of heated atoms in thermodynamics~\cite{kirkpatrick1983optimization}. In the ALVNS-SA algorithm, the SA mechanism is employed following the evaluation of the repair scenario. Despite potentially being less hazardous compared to previously tested scenarios, there remains a probability of accepting the repair outcome. By accepting less dangerous scenarios, the algorithm can break free from local collision scenarios and persist in the search process. This enables the algorithm to explore diverse search spaces and increase the likelihood of discovering more perilous scenarios.

\begin{figure}[htbp]
    \centering
    \includegraphics[width=0.3\textwidth]{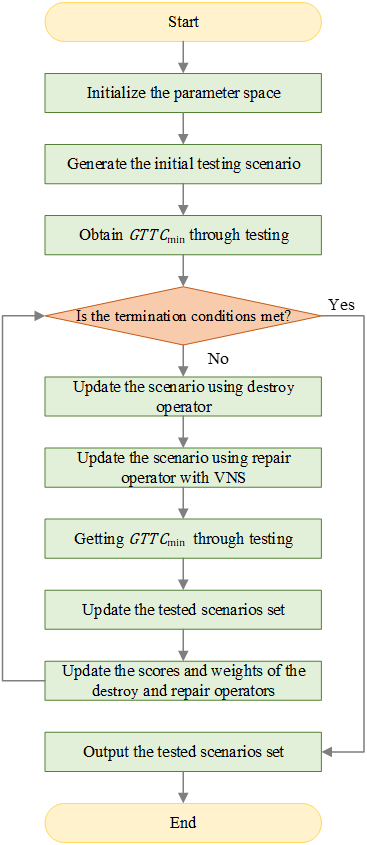}
    \caption{Crash-based counterfactual simulation}
    \label{fig:ALVNS-SA}
\end{figure}

The framework of ALVNS-SA is shown in Fig.~\ref{fig:ALVNS-SA} and the pseudo-code is presented in Algorithm~1, while Table~\ref{tab:alvns_parameters} provides detailed descriptions of the algorithm's parameters.

\begin{table}[htbp]
\caption{Description of ALVNS-SA Parameters}
\label{tab:alvns_parameters}
\centering
\renewcommand{\arraystretch}{1.2}
\begin{tabular}{|c|p{5.1cm}|}
\hline
\textbf{Parameter} & \textbf{Description} \\
\hline
$D$ & Parameter space \\
$x^i$ & $i$th parameter \\
$\gamma^i$ & Discrete step size of the $i$th parameter \\
$x^i_{range}$ & Range of the $i$th parameter, $x^i_{max} - x^i_{min}$ \\
$\mathcal{N}^i_j$ & $j$th neighborhood of the $i$th parameter \\
$R^-_i$ & $i$th destroy operator \\
$R^+_i$ & $i$th repair operator \\
$w^d_i$ & Weight of $R^-_i$ \\
$w^r_i$ & Weight of $R^+_i$ \\
$u^d_i$ & Number of uses of $R^-_i$ \\
$u^r_i$ & Number of uses of $R^+_i$ \\
$s^d_i$ & Cumulative score of $R^-_i$ \\
$s^r_i$ & Cumulative score of $R^+_i$ \\
$T_b$ & SA initial temperature \\
$T_e$ & SA termination temperature \\
$T_c$ & SA current temperature \\
$\alpha$ & Cooling rate \\
\hline
\end{tabular}
\end{table}

\begin{algorithm}[htbp]
\caption{ALVNS-SA}
\label{alg:alvns-sa}
\KwIn{$D$, $R_i^-$, $R_i^+$, $T_b$, $T_e$, $T_c$, $\alpha$, $w_i^d$, $w_i^r$, $s_i^d$, $s_i^r$, termination condition $\tau$}
\KwOut{High risk scenario set $\Omega^*$}

$s_0 \leftarrow$ initial scenario\;
$s \leftarrow s_0$, $T_c \leftarrow T_b$, $\Omega^* \leftarrow \emptyset$\;

\While{$\tau$ has not been reached}{
    \While{$T_c > T_e$}{
        Generate the destroy scenario $s_d$\;
        Generate the repair scenario $s_r$ based on VNS\;
        $s' \leftarrow s_r$\;
        
        \eIf{$f(s') < f(s)$}{
            $\Omega^* \leftarrow \Omega^* + s'$\;
            $s \leftarrow s'$\;
        }{
            \If{$f(s') > f(s)$ and meets the acceptance criteria for SA}{
                $\Omega^* \leftarrow \Omega^* + s'$\;
                $s \leftarrow s'$\;
            }
        }
        $T_c \leftarrow \alpha \cdot T_c$\;
        Update the scores and weights of the destroy and repair operators\;
    }
    Reset the temperature: $T_c = T_b$\;
}
\end{algorithm}

ALVNS-SA is an iterative algorithm that uses gradual improvement of initial testing scenarios for selection. This algorithm gradually tests different levels of safety-critical scenarios within the scenario space. Compared to random testing, ALVNS-SA can automatically test more safety-critical scenarios, thereby accelerating the capability testing of AVs. By continually improving and testing scenarios with varying levels of risk, ALVNS-SA helps ensure that autonomous driving systems can operate correctly in various complex and dangerous situations. In this way, the algorithm plays an important role in enhancing the safety performance of AVs.

\subsubsection{Destroy Operator for ALVNS-SA}
For this study, eight destroy operators are defined as follows:

\begin{equation}
\begin{aligned}
R_1 &= v_1 - \xi, &\quad R_2 &= v_1 + \xi, \\
R_3 &= v_2 - \xi, &\quad R_4 &= v_2 + \xi, \\
R_5 &= d - \xi, &\quad R_6 &= d + \xi, \\
R_7 &= a_1 - \xi, &\quad R_8 &= a_1 + \xi.
\end{aligned}
\end{equation}

The initial scores of $R_1$, $R_2$, $R_3$, and $R_7$ are set to 1.5, and others are set to 1. All destroy operators share the same initial weight of 1.

The value $\xi$ is a random number within the destruction step length. When the number of consecutive rejections exceeds the threshold, $\xi$ is adaptively revised according to $GTTC_{min}$ and the number of iterations:

\begin{equation}
\xi \in
\begin{cases}
[0, 0.1 \cdot x_{\text{range}}^i], & \text{if crash} \\
[0, 0.2 \cdot x_{\text{range}}^i], & \text{if } 0.5 \leq GTTC_{min} \\
[0, 0.3 \cdot x_{\text{range}}^i], & \text{if } 1 \leq GTTC_{min} \\
[0, 0.8 \cdot x_{\text{range}}^i], & \text{if } 1 \leq GTTC_{min} \leq 2 \\
\left[0, \left(0.8 - 0.4 \cdot \frac{it}{it_{\max}}\right) \cdot x_{\text{range}}^i \right], & \text{if } 2 < GTTC_{min}
\end{cases}
\end{equation}

\begin{algorithm}[htbp]
\caption{VNS}
\KwIn{
    $L = \max_{i=1,2,3,4} \left( \frac{x^i_{\text{range}}}{\gamma^i} \right)$
}
\For{$j$ in range$(1, L)$}{
    Generate $s_d$'s $j$th neighborhood\;
    $\mathcal{N}^j_{s_d} = \left[ s^i_d - j \cdot \gamma^i, \quad s^i_d + j \cdot \gamma^i \right]$\;
    Generate all untested scenarios set $\Omega_x$ within $\mathcal{N}^j_{s_d}$\;
    $\Omega_{\text{untested}} = \Omega_x - \Omega_{\text{tested}}$\;
    \If{$\text{size}(\Omega_{\text{untested}}) \geq 2$}{
        Compute the distance $d_l$ between each $s^l_{\text{untested}} \in \Omega_{\text{untested}}$ and $s_d$\;
        Sort all $s_{\text{untested}} \in \Omega_{\text{untested}}$ in ascending order according to $d_l$\;
        Select the two scenarios $(s_1, s_2)$ from $\Omega_{\text{untested}}$ closest to scenario $s_d$\;
        Use roulette wheel selection to choose $R^+_{k}$ based on $w^r_k, \; (k=1,2)$\;
        $s_r = s_k \oplus R^+_k$\;
    }
    \ElseIf{$\text{size}(\Omega_{\text{untested}}) = 1$}{
        Use roulette wheel selection to choose $R^+_{k}$ based on $w^r_k, \; (k=1,2)$\;
        $s_r = s_1 \oplus R^+_k$\;
    }
}
\end{algorithm}

\subsubsection{ Iterative Approach for ALVNS-SA}
\begin{itemize}
    \item \textbf{Destroy stage:} Based on the weights of the destroy operators and using the roulette wheel selection method, select a destroy operator to apply on the current scenario and generate a destructed solution.
    
    \item \textbf{Repair stage:} Based on the weights of the repair operators, using ANS and the roulette wheel selection method, select a repair operator to apply on the destructed scenario to obtain a new scenario. The new scenario is then input into a simulation model to obtain its objective function value, which is the minimization of $GTTC_{min}$.
    
    \item \textbf{Acceptance criteria and SA:} Based on the value of the objective function, $\Delta GTTC_{min} \leq 0$, determine whether to accept the new solution as the current scenario. For instance, if the change in the objective function $\Delta GTTC_{min} \leq 0$, always accept the new scenario. Otherwise, enter the SA process: Accept the new scenario with a probability 
    \[
    P = \exp\left(\frac{-\Delta GTTC_{min}}{T_c}\right)
    \]
    where $T_c$ is the current temperature in simulated annealing. Update the current temperature based on the cooling rate.
    
    \item \textbf{Updates:} Based on the nature of the algorithm, and considering the usage of operators, objective function, and acceptance of scenarios, update the scores, usage count, and weights of the destroy and repair operators. The weight calculation can be performed as follows:
    
    \[
    w = 
    \begin{cases}
    w, & \text{if } u = 0 \\
    (1 - \rho)w + \rho \frac{s}{u}, & \text{if } u > 0
    \end{cases}
    \]
    
    where $w$ is the operator weight, $s$ is the operator score, $u$ is the number of times the operator has been used, and $\rho$ is the weight update coefficient.
    
    The new score $s'$ is calculated as:
    
    \[
    s' = s + \vartheta
    \]
\end{itemize}

5) Scenarios storage: store tested scenarios into a collection of tested scenarios to avoid repeated testing.

\begin{table}[htbp]
\centering
\caption{Assignment Rule for $\vartheta$ under Crash ($GTTC'_{min} < GTTC_{min}$)}
\label{tab:theta_crash}
\renewcommand{\arraystretch}{1.3}
\begin{tabular}{|c|c|}
\hline
\textbf{Condition} & \textbf{$\vartheta$ Value} \\
\hline
$GTTC'_{min} \leq 0.5$ & 2.6 \\
$0.5 < GTTC'_{min} \leq 1.0$ & 2.2 \\
$1.0 < GTTC'_{min} \leq 2.0$ & 1.8 \\
$GTTC'_{min} > 2.0$ & 0.2 \\
\hline
\end{tabular}
\end{table}

\begin{table}[htbp]
\centering
\caption{Assignment Rule for $\vartheta$ under No-crash ($GTTC'_{min} \geq GTTC_{min}$)}
\label{tab:theta_nocrash}
\renewcommand{\arraystretch}{1.3}
\begin{tabular}{|c|c|c|}
\hline
\textbf{Condition} & \makecell{\textbf{Accepted} \\ \textbf{in SA}} & \makecell{\textbf{Rejected} \\ \textbf{in SA}} \\
\hline
$GTTC'_{min} \leq 0.5$ & 2.0 & 1.8 \\
$0.5 < GTTC'_{min} \leq 1.0$ & 1.6 & 1.4 \\
$1.0 < GTTC'_{min} \leq 2.0$ & 1.2 & 1.0 \\
$GTTC'_{min} > 2.0$ & 0.1 & 0 \\
\hline
\end{tabular}
\end{table}

\section{EXPERIMENTS AND ALGORITHM COMPARISON RESULTS}

All experiments and scenario analyses were conducted on Apollo 7.0 and SVL Simulator. The experimental equipment used an operating system of Ubuntu 18.04, with an NVIDIA GeForce RTX 2080 Ti GPU and an eight-core Intel Core i9-9900K @ 3.60GHz processor. All algorithms were coded in Python 3.8.

\subsection{ Optimization Testing Based on ALVNS-SA}

\begin{figure}[htbp]
    \centering
    \includegraphics[width=0.48\textwidth]{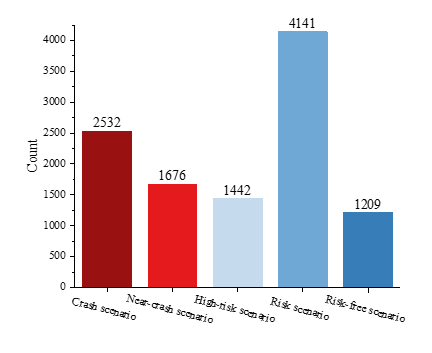}
    \caption{Distribution of test scenarios based on ALVNS-SA}
    \label{fig:Distribution of test scenarios based on ALVNS-SA}
\end{figure}

The percentage of each type of scenario in ALVNS-SA out of 11,000 tests is depicted in Fig. \ref{fig:Distribution of test scenarios based on ALVNS-SA}. Based on the experimental results presented in Fig. \ref{fig:Distribution of test scenarios based on ALVNS-SA}, it is evident that ALVNS-SA demonstrates remarkable effectiveness in identifying safety-critical scenarios and facilitating accelerated testing. The test samples revealed that these scenarios account for 23.02\%, 14.46\%, 12.08\%, and 34.44\%, respectively. The risk-free scenarios only represent 16.0\%.
The experimental results signify the effectiveness of ALVNS-SA as it can identify safety-critical scenarios during the testing process and focusing on situations that may have potential issues and risks. ALVNS-SA, through conducting over 11,000 tests, examined over 80\% of various types of safety-critical scenarios, providing evidence for its effectiveness in searching for safety-critical scenarios and completing accelerated testing.

They provide a foundation for further optimization of testing strategies, enhancement of system performance, and mitigation of potential risks. The application of ALVNS in testing scenarios could significantly contribute to the development and improvement of various systems and processes.

In conclusion, the thorough analysis of the experimental results from Fig. \ref{fig:Distribution of test scenarios based on ALVNS-SA} and Table 4 unequivocally demonstrates the remarkable effectiveness of ALVNS in searching for critical scenarios and accomplishing accelerated testing. The implications of these findings extend beyond academia, with practical implications for optimizing testing strategies and improving system performance while reducing risks.

\subsection{Algorithm comparison}

In order to contrast the advantages of ALVNS-SA in optimization testing, this study also compares ALNS-SA without VNS, GA and parameter random combination testing. According to the problem characteristics of this study, the design of GA is presented as Algorithm~3. The parameter settings for the genetic algorithm are as follows: $M = 100$, $P_c = 0.75$, $P_m = 0.05$, $T = 1500$, $\tau = 11000$. For parameter-random combination testing, randomly selecting parameter values and conducting 11{,}000 tests.

\begin{algorithm}[htbp]
\caption{GA for Accelerated Testing}
\KwIn{Parameter space $D$, fitness function $F$, initial population $P_0$, population size $M$, crossover probability $P_c$, mutation probability $P_m$, maximum number of iterations $T$, terminal condition $\tau$}
\KwOut{Scenario set $\Omega^*$}
\textbf{Initial population:} $P_0$, let $\Omega^* = \emptyset$, $t = 0$\;
Calculate the fitness $F^i$ of each individual\;
\While{$\tau$ has not been reached}{
    \While{$t \leq T$}{
        $\Theta^{t+1} \leftarrow \emptyset$\;
        \For{$i = 1$ \KwTo $M$}{
            Select individual $P^t_i$ as one parent\;
            Use fitness $F^t$ for roulette wheel selection to choose another individual $P^t_k$\;
            \If{$l_1 = \text{random}(0,1) < P_c$}{
                Generate offspring individuals $P^t_{c1}$ and $P^t_{c2}$\;
                $\Theta^{t+1} \leftarrow [P^t_{c1}, P^t_{c2}]$\;
            }
            \If{$l_1 = \text{random}(0,1) < P_m$}{
                Generate new offspring individuals $P^t_{m1}$ and $P^t_{m2}$\;
                $\Theta^{t+1} \leftarrow [P^t_{m1}, P^t_{m2}]$\;
            }
        }
        \ForEach{$P^i \in \Theta^{t+1}$}{
            \If{$P^i \in \Omega^*$}{
                Replace $P^i$ by $P^i_\text{nearest} \in \Omega_\text{untested}$\;
            }
        }
        $\Omega^* = \Omega^* + \Theta^{t+1}$\;
        Select the $M$ individuals with the lowest fitness as the new population $P^{t+1}$\;
        $t = t + 1$\;
    }
}
\end{algorithm}

ALNS-SA is a heuristic search algorithm in which VNS is removed from ALVNS-SA. The rest of the operational parameters are the same as ALVNS-SA. We aim to determine the advantages and improvements of the VNS strategy through ALNS-SA and ALVNS-SA comparison experiments.

In order to evaluate the performance of different algorithms, we utilize the metric of scenario coverage rate ($\Lambda$) and the proportion ($P$) of tested scenarios. The crash scenario proportion $P_\text{crash}(i)$ for algorithm $i$ is calculated as follows:

\begin{equation}
P_\text{crash}(i) = \frac{\Omega_i^\text{crash}}{\Omega_i^\text{crash} \cup \Omega_i^\text{near-crash} \cup \Omega_i^\text{high-risk} \cup \Omega_i^\text{risk} \cup \Omega_i^\text{risk-free}}
\end{equation}

$P_\text{near-crash}(i)$, $P_\text{high-risk}(i)$, $P_\text{risk}(i)$ and $P_\text{risk-free}(i)$ can be calculated by the same method.

The crash scenario coverage rate $\Lambda_\text{crash}(i)$ for algorithm $i$ is calculated as follows:

\begin{equation}
\Lambda_\text{crash}(i) = \frac{\Omega_i^\text{crash}}{\Omega_\text{ALVNS-SA}^\text{crash} \cup \Omega_\text{ALNS-SA}^\text{crash} \cup \Omega_\text{GA}^\text{crash} \cup \Omega_\text{Random}^\text{crash}}
\end{equation}

The same method can be used to calculate $\Lambda_\text{near-crash}(i)$, $\Lambda_\text{high-risk}(i)$, $\Lambda_\text{risk}(i)$, and $\Lambda_\text{risk-free}(i)$ separately. The comparison results are shown in Table~\ref{tab:comparison-methods}.

Table~\ref{tab:comparison-methods} illustrates the tested scenarios from different algorithms. It is apparent that ALVNS-SA covers a higher proportion of safety-critical scenarios compared to the other three testing approaches.

\begin{table}[htbp]
\centering
\caption{Comparison of Various Testing Methods}
\label{tab:comparison-methods}
\begin{tabular}{lcc}
\toprule
\textbf{Category} & \textbf{Metric} & \textbf{Value (\%)} \\
\midrule
\multirow{2}{*}{Crash scenario} 
  & $P$ (ALVNS-SA) & 23.02 \\
  & $\Lambda$ (ALVNS-SA) & 96.83 \\
  & $P$ (ALNS-SA) & 13.61 \\
  & $\Lambda$ (ALNS-SA) & 78.09 \\
  & $P$ (GA) & 13.47 \\
  & $\Lambda$ (GA) & 77.25 \\
  & $P$ (Random) & 4.37 \\
  & $\Lambda$ (Random) & 25.05 \\
\midrule
\multirow{2}{*}{Near-crash scenario} 
  & $P$ (ALVNS-SA) & 14.46 \\
  & $\Lambda$ (ALVNS-SA) & 92.07 \\
  & $P$ (ALNS-SA) & 8.91 \\
  & $\Lambda$ (ALNS-SA) & 77.37 \\
  & $P$ (GA) & 9.54 \\
  & $\Lambda$ (GA) & 82.81 \\
  & $P$ (Random) & 2.97 \\
  & $\Lambda$ (Random) & 25.75 \\
\midrule
\multirow{2}{*}{High-risk scenario} 
  & $P$ (ALVNS-SA) & 12.08 \\
  & $\Lambda$ (ALVNS-SA) & 84.38 \\
  & $P$ (ALNS-SA) & 7.41 \\
  & $\Lambda$ (ALNS-SA) & 70.54 \\
  & $P$ (GA) & 7.74 \\
  & $\Lambda$ (GA) & 73.71 \\
  & $P$ (Random) & 2.49 \\
  & $\Lambda$ (Random) & 23.75 \\
\midrule
\multirow{2}{*}{Risk scenario} 
  & $P$ (ALVNS-SA) & 34.44 \\
  & $\Lambda$ (ALVNS-SA) & 71.65 \\
  & $P$ (ALNS-SA) & 21.32 \\
  & $\Lambda$ (ALNS-SA) & 60.49 \\
  & $P$ (GA) & 26.19 \\
  & $\Lambda$ (GA) & 74.30 \\
  & $P$ (Random) & 8.85 \\
  & $\Lambda$ (Random) & 25.10 \\
\midrule
\multirow{2}{*}{Risk-free scenario} 
  & $P$ (ALVNS-SA) & 16.00 \\
  & $\Lambda$ (ALVNS-SA) & 3.57 \\
  & $P$ (ALNS-SA) & 48.75 \\
  & $\Lambda$ (ALNS-SA) & 14.84 \\
  & $P$ (GA) & 43.07 \\
  & $\Lambda$ (GA) & 13.11 \\
  & $P$ (Random) & 81.33 \\
  & $\Lambda$ (Random) & 24.76 \\
\bottomrule
\end{tabular}
\end{table}

According to the results in Table~\ref{tab:comparison-methods}. The ALVNS-SA revealed remarkable coverage rates: the crash scenario achieved an impressive coverage rate of 96.83\%, the near-crash scenario reached 92.07\% coverage, the high-risk scenario achieved 84.38\% coverage, and the risk scenario obtained a coverage rate of 71.65\%. ALVNS-SA is significantly better than the other three testing methods in searching for safety-critical scenarios.

\section{Discussion}

One of the key applications of the proposed ALVNS-SA algorithm is to accelerate the testing process for AVs. Due to the vastness of the parameter space in scenario-based testing~\cite{wang2022safety}, identifying appropriate combinations for constructing concrete scenarios presents a substantial challenge.

Logical scenarios derived from naturalistic driving data often involve low-conflict situations. In contrast, real-world crash data inherently represent safety-critical scenarios and can be leveraged to extract typical logical scenarios that reflect high-risk conditions~\cite{zhou2024would,sui2021emergency,abdessalem2018testing,nilsson2018definition}. These typical logical scenarios encapsulate key environmental and behavioral factors that are more likely to lead to crashes, thereby enhancing the relevance and efficiency of AV safety testing. Compared with randomly generated scenarios, testing based on these critical scenarios helps reduce redundancy, focus on high-impact situations, and optimize the use of computational resources.

However, even for logical scenarios extracted from crash data, the corresponding parameter space remains vast. For instance, in this study, a two-vehicle rear-end collision scenario defined by just four parameters yields over 60{,}000 possible concrete scenarios. When environmental and behavioral parameters are also considered, the parameter space expands exponentially. Therefore, it becomes essential to employ optimization techniques that bridge logical and concrete scenarios, enabling the efficient identification of high-risk cases.

By integrating ALVNS-SA into the testing workflow, the search for safety-critical concrete scenarios is significantly accelerated. The algorithm effectively navigates the high-dimensional parameter space to generate diverse and high-risk scenarios, thereby increasing test coverage and enabling the discovery of edge-case behaviors that may be missed by random or manual approaches.

Compared to manual or random selection, ALVNS-SA intelligently selects optimal scenario combinations tailored to specific testing objectives. This not only improves testing efficiency and reduces time and cost but also helps identify performance bottlenecks and interaction effects in AV systems under different scenario configurations~\cite{wang2022safety}. As such, ALVNS-SA serves as a valuable tool for advancing the robustness and reliability of AV evaluation frameworks.

\section{Conclusion}

This study presents an optimization-based testing method using the ALVNS-SA algorithm to accelerate the identification of safety-critical scenarios from real-world crash data. By designing eight destroy and two repair operators,enhanced with VNS,the proposed method effectively explores the high-dimensional parameter space of logical scenarios.

Experimental results show that ALVNS-SA significantly outperforms G) and random testing in terms of crash scenario coverage and overall efficiency. The algorithm successfully identified over 80\% of critical scenario types across 11,000 test cases, demonstrating its robustness and applicability to black-box AV testing.

Despite these promising results, this study is limited to rear-end collision scenarios and uses $GTTC_{min}$ as the sole safety metric, without considering scenario complexity or environmental factors. Future work will extend the method to a wider range of scenarios and adopt more comprehensive evaluation criteria.

% or




%\bibliographystyle{IEEEtran}
\bibliographystyle{IEEEtran}

\bibliography{IEEEexample}

\begin{thebibliography}{10}
\providecommand{\url}[1]{#1}
\csname url@samestyle\endcsname
\providecommand{\newblock}{\relax}
\providecommand{\bibinfo}[2]{#2}
\providecommand{\BIBentrySTDinterwordspacing}{\spaceskip=0pt\relax}
\providecommand{\BIBentryALTinterwordstretchfactor}{4}
\providecommand{\BIBentryALTinterwordspacing}{\spaceskip=\fontdimen2\font plus
\BIBentryALTinterwordstretchfactor\fontdimen3\font minus \fontdimen4\font\relax}
\providecommand{\BIBforeignlanguage}[2]{{%
\expandafter\ifx\csname l@#1\endcsname\relax
\typeout{** WARNING: IEEEtran.bst: No hyphenation pattern has been}%
\typeout{** loaded for the language `#1'. Using the pattern for}%
\typeout{** the default language instead.}%
\else
\language=\csname l@#1\endcsname
\fi
#2}}
\providecommand{\BIBdecl}{\relax}
\BIBdecl

\bibitem{xu2019statistical}
C.~Xu, Z.~Ding, C.~Wang, and Z.~Li, ``Statistical analysis of the patterns and characteristics of connected and autonomous vehicle involved crashes,'' \emph{Journal of Safety Research}, vol.~71, pp. 41--47, Dec 2019.

\bibitem{huang2024pre}
H.~Huang, X.~Huang, R.~Zhou, H.~Zhou, J.~J. Lee, and X.~Cen, ``Pre-crash scenarios for safety testing of autonomous vehicles: A clustering method for in-depth crash data,'' \emph{Accident Analysis \& Prevention}, vol. 203, p. 107616, 2024.

\bibitem{zhou2024evaluating}
R.~Zhou, Z.~Lin, G.~Zhang, H.~Huang, H.~Zhou, and J.~Chen, ``Evaluating autonomous vehicle safety performance through analysis of pre-crash trajectories of powered two-wheelers,'' \emph{IEEE Transactions on Intelligent Transportation Systems}, vol.~25, no.~10, pp. 13\,560--13\,572, 2024.

\bibitem{jin2024analysis}
J.~Jin, H.~Huang, R.~Zhou, J.~Chen, and P.~Liu, ``Analysis of dynamic determinants of vehicles involved in crash affecting severity based on in-depth crash data,'' \emph{Traffic injury prevention}, vol.~25, no.~3, pp. 537--543, 2024.

\bibitem{zhou2024would}
R.~Zhou, G.~Zhang, H.~Huang, Z.~Wei, H.~Zhou, J.~Jin, F.~Chang, and J.~Chen, ``How would autonomous vehicles behave in real-world crash scenarios?'' \emph{Accident Analysis \& Prevention}, vol. 202, p. 107572, 2024.

\bibitem{li2025multidimensional}
S.~Li, R.~Zhou, and H.~Huang, ``Multidimensional evaluation of autonomous driving test scenarios based on ahp-ewn-topsis models,'' \emph{Automotive Innovation}, pp. 1--15, 2025.

\bibitem{zhou2022testing}
R.~Zhou, Z.~Lin, X.~Huang, J.~Peng, and H.~Huang, ``Testing scenarios construction for connected and automated vehicles based on dynamic trajectory clustering method,'' in \emph{2022 IEEE 25th International Conference on Intelligent Transportation Systems (ITSC)}.\hskip 1em plus 0.5em minus 0.4em\relax IEEE, 2022, pp. 3304--3308.

\bibitem{lenard2014pedestrian}
J.~Lenard, ``Typical pedestrian accident scenarios for the development of autonomous emergency braking test protocols,'' \emph{Accident Analysis \& Prevention}, p.~8, 2014.

\bibitem{nitsche2017precrash}
P.~Nitsche, P.~Thomas, R.~Stuetz, and R.~Welsh, ``Pre-crash scenarios at road junctions: A clustering method for car crash data,'' \emph{Accident Analysis \& Prevention}, vol. 107, pp. 137--151, Oct 2017.

\bibitem{sui2021emergency}
B.~Sui, N.~Lubbe, and J.~Bärgman, ``Evaluating automated emergency braking performance in simulated car-to-two-wheeler crashes in china: A comparison between c-ncap tests and in-depth crash data,'' \emph{Accident Analysis \& Prevention}, vol. 159, p. 106229, Sep 2021.

\bibitem{wang2022autonomous}
X.~Wang \emph{et~al.}, ``Autonomous driving testing scenario generation based on in-depth vehicle-to-powered two-wheeler crash data in china,'' \emph{Accident Analysis \& Prevention}, vol. 176, p. 106812, Oct 2022.

\bibitem{zhou2023precrash}
R.~Zhou, H.~Huang, J.~Lee, X.~Huang, J.~Chen, and H.~Zhou, ``Identifying typical pre-crash scenarios based on in-depth crash data with deep embedded clustering for autonomous vehicle safety testing,'' \emph{Accident Analysis \& Prevention}, vol. 191, p. 107218, Oct 2023.

\bibitem{zhang2025high}
G.~Zhang, H.~Huang, R.~Zhou, S.~Li, and J.~Bian, ``High-risk trajectories generation for safety testing of autonomous vehicles based on in-depth crash data,'' \emph{IEEE Transactions on Intelligent Transportation Systems}, 2025.

\bibitem{huai2023scenorita}
Y.~Huai, S.~Almanee, Y.~Chen, X.~Wu, Q.~A. Chen, and J.~Garcia, ``scenorita: Generating diverse, fully-mutable, test scenarios for autonomous vehicle planning,'' \emph{IEEE Transactions on Software Engineering}, pp. 1--21, 2023.

\bibitem{abdessalem2018testing}
R.~B. Abdessalem, S.~Nejati, L.~C. Briand, and T.~Stifter, ``Testing vision-based control systems using learnable evolutionary algorithms,'' in \emph{Proc. Int. Conf. Software Engineering}.\hskip 1em plus 0.5em minus 0.4em\relax Gothenburg, Sweden: ACM, 2018, pp. 1016--1026.

\bibitem{beglerovic2017model}
H.~Beglerovic, A.~Ravi, N.~Wikström, H.-M. Koegeler, A.~Leitner, and J.~Holzinger, ``Model-based safety validation of the automated driving function highway pilot,'' in \emph{8th International Munich Chassis Symposium 2017: chassis. tech plus}.\hskip 1em plus 0.5em minus 0.4em\relax Springer, 2017, pp. 309--329.

\bibitem{corso2019adaptive}
A.~Corso, P.~Du, K.~R. Driggs-Campbell, and M.~J. Kochenderfer, ``Adaptive stress testing with reward augmentation for autonomous vehicle validation,'' in \emph{2019 IEEE Intelligent Transportation Systems Conference (ITSC)}, 2019, pp. 163--168.

\bibitem{menzel2018scenarios}
T.~Menzel, G.~Bagschik, and M.~Maurer, ``Scenarios for development, test and validation of automated vehicles,'' in \emph{2018 IEEE Intelligent Vehicles Symposium (IV)}, 2018, pp. 1821--1827.

\bibitem{wang2022gradient}
Y.~Wang, R.~Yu, S.~Qiu, J.~Sun, and H.~Farah, ``Safety-performance-boundary identification of highly automated vehicles: A surrogate-model-based gradient descent searching approach,'' \emph{IEEE Transactions on Intelligent Transportation Systems}, vol.~23, no.~12, pp. 23\,809--23\,820, Dec 2022.

\bibitem{huang2022framework}
X.~Huang, X.~Cen, M.~Cai, and R.~Zhou, ``A framework to analyze function domains of autonomous transportation systems based on text analysis,'' \emph{Mathematics}, vol.~11, no.~1, p. 158, 2022.

\bibitem{huang2022functional}
X.~Huang, H.~Huang, X.~Cen, M.~Cai, R.~Zhou, and Y.~Li, ``Functional domains clustering of autonomous transportation systems based on latent dirichlet allocation,'' in \emph{CICTP 2022}, 2022, pp. 367--377.

\bibitem{arief2021deep}
M.~Arief \emph{et~al.}, ``Deep probabilistic accelerated evaluation: A robust certifiable rare-event simulation methodology for black-box safety-critical systems,'' in \emph{24th International Conference on Artificial Intelligence and Statistics (AISTATS 2021)}, 2021, pp. 595--603.

\bibitem{xu2018accelerated}
Y.~Xu, Y.~Zou, and J.~Sun, ``Accelerated testing for automated vehicles safety evaluation in cut-in scenarios based on importance sampling, genetic algorithm and simulation applications,'' \emph{Journal of Intelligent \& Connected Vehicles}, vol.~1, no.~1, pp. 28--38, 2018.

\bibitem{majzik2019towards}
I.~Majzik \emph{et~al.}, ``Towards system-level testing with coverage guarantees for autonomous vehicles,'' in \emph{2019 Proc. ACM/IEEE Int. Conf. Model-Driven Eng. Lang. Syst. (MODELS)}, 2019, pp. 89--94.

\bibitem{zhao2017accelerated}
D.~Zhao \emph{et~al.}, ``Accelerated evaluation of automated vehicles safety in lane-change scenarios based on importance sampling techniques,'' \emph{IEEE Transactions on Intelligent Transportation Systems}, vol.~18, no.~3, pp. 595--607, Mar 2017.

\bibitem{sun2022scenario}
J.~Sun, H.~Zhang, H.~Zhou, R.~Yu, and Y.~Tian, ``Scenario-based test automation for highly automated vehicles: A review and paving the way for systematic safety assurance,'' \emph{IEEE Transactions on Intelligent Transportation Systems}, vol.~23, no.~9, pp. 14\,088--14\,103, Sep 2022.

\bibitem{batsch2023taxonomy}
F.~Batsch, S.~Kanarachos, M.~Cheah, R.~Ponticelli, and M.~Blundell, ``A taxonomy of validation strategies to ensure the safe operation of highly-automated vehicles,'' \emph{Journal of Intelligent Transportation Systems: Technology, Planning, and Operations}, vol.~27, no.~1, pp. 143--153, 2023.

\bibitem{gambi2019automatically}
A.~Gambi, M.~Mueller, and G.~Fraser, ``Automatically testing self-driving cars with search-based procedural content generation,'' in \emph{ISSTA - Proc. ACM SIGSOFT Int. Symp. Softw. Test. Anal.}, Beijing, China, 2019, pp. 318--328.

\bibitem{zhou2025crash}
R.~Zhou, H.~Huang, G.~Zhang, H.~Zhou, and J.~Bian, ``Crash-based safety testing of autonomous vehicles: Insights from generating safety-critical scenarios based on in-depth crash data,'' \emph{IEEE Transactions on Intelligent Transportation Systems}, 2025.

\bibitem{cai2020dynamic}
J.~Cai, R.~Zhou, and D.~Lei, ``Dynamic shuffled frog-leaping algorithm for distributed hybrid flow shop scheduling with multiprocessor tasks,'' \emph{Engineering Applications of Artificial Intelligence}, vol.~90, p. 103540, 2020.

\bibitem{zhou2019multi}
R.~Zhou, D.~Lei, and X.~Zhou, ``Multi-objective energy-efficient interval scheduling in hybrid flow shop using imperialist competitive algorithm,'' \emph{IEEE access}, vol.~7, pp. 85\,029--85\,041, 2019.

\bibitem{cai2020fuzzy}
J.~Cai, R.~Zhou, and D.~Lei, ``Fuzzy distributed two-stage hybrid flow shop scheduling problem with setup time: collaborative variable search,'' \emph{Journal of intelligent \& fuzzy systems}, vol.~38, no.~3, pp. 3189--3199, 2020.

\bibitem{cao2019typical}
Y.~Cao \emph{et~al.}, ``Typical pre-crash scenarios reconstruction for two-wheelers and passenger vehicles and its application in parameter optimization of aeb system based on nais database,'' \url{https://trid.trb.org/view/1758949}, 2019, presented at the 26th International Technical Conference on the Enhanced Safety of Vehicles (ESV), Highway Traffic Safety Administration.

\bibitem{nilsson2018definition}
D.~Nilsson, M.~Lindman, T.~Victor, and M.~Dozza, ``Definition of run-off-road crash clusters—for safety benefit estimation and driver-assistance development,'' \emph{Accident Analysis \& Prevention}, vol. 113, pp. 97--105, Apr 2018.

\bibitem{pan2021study}
D.~Pan, Y.~Han, Q.~Jin, H.~Wu, and H.~Huang, ``Study of typical electric two-wheelers pre-crash scenarios using k-medoids clustering methodology based on video recordings in china,'' \emph{Accident Analysis \& Prevention}, vol. 160, p. 106320, Sep 2021.

\bibitem{zhu2021hazardous}
B.~Zhu, P.~Zhang, J.~Zhao, and W.~Deng, ``Hazardous scenario enhanced generation for automated vehicle testing based on optimization searching method,'' \emph{IEEE Transactions on Intelligent Transportation Systems}, vol.~23, no.~7, pp. 7321--7331, 2021.

\bibitem{zhao2025cradle}
J.~Zhao, W.~Li, B.~Zhu, P.~Zhang, Y.~Huang, and R.~Tang, ``Cradle: An accident scenario generation method based on scenario knowledge graph considering accident causation,'' \emph{IEEE Transactions on Software Engineering}, 2025.

\bibitem{bian2025search}
J.~Bian, H.~Huang, Q.~Yu, and R.~Zhou, ``Search-to-crash: Generating safety-critical scenarios from in-depth crash data for testing autonomous vehicles,'' \emph{Energy}, p. 137174, 2025.

\bibitem{wei2025generating}
Z.~Wei, J.~Bian, H.~Huang, R.~Zhou, and H.~Zhou, ``Generating risky and realistic scenarios for autonomous vehicle tests involving powered two-wheelers: A novel reinforcement learning framework,'' \emph{Accident Analysis \& Prevention}, vol. 218, p. 108038, 2025.

\bibitem{beglerovic2017testing}
H.~Beglerovic, M.~Stolz, and M.~Horn, ``Testing of autonomous vehicles using surrogate models and stochastic optimization,'' in \emph{2017 IEEE Intelligent Transportation Systems Conference (ITSC)}, 2017, pp. 1--6.

\bibitem{mullins2018adaptive}
G.~Mullins, ``Adaptive sampling methods for testing autonomous systems,'' 2018.

\bibitem{tuncali2016utilizing}
C.~E. Tuncali, T.~P. Pavlic, and G.~Fainekos, ``Utilizing s-taliro as an automatic test generation framework for autonomous vehicles,'' in \emph{IEEE Conf. Intell. Transport Syst. Proc. ITSC}, 2016, pp. 1470--1475.

\bibitem{tuncali2017functional}
C.~E. Tuncali, S.~Yaghoubi, T.~P. Pavlic, and G.~Fainekos, ``Functional gradient descent optimization for automatic test case generation for vehicle controllers,'' in \emph{2017 13th IEEE Int. Conf. Autom. Sci. Eng. (CASE)}, 2017, pp. 1059--1064.

\bibitem{tuncali2019requirements}
C.~E. Tuncali, G.~Fainekos, D.~Prokhorov, H.~Ito, and J.~Kapinski, ``Requirements-driven test generation for autonomous vehicles with machine-learning components,'' \emph{IEEE Transactions on Intelligent Vehicles}, vol.~5, no.~2, pp. 265--280, 2019.

\bibitem{calo2020generating}
A.~Calò, P.~Arcaini, S.~Ali, F.~Hauer, and F.~Ishikawa, ``Generating avoidable collision scenarios for testing autonomous driving systems,'' in \emph{2020 Proc. IEEE Int. Conf. Softw. Test., Verif. Valid. (ICST)}, 2020, pp. 375--386.

\bibitem{batsch2019performance}
F.~Batsch, A.~Daneshkhah, M.~Cheah, S.~Kanarachos, and A.~Baxendale, ``Performance-boundary identification for the evaluation of automated vehicles using gaussian process classification,'' in \emph{2019 IEEE Intelligent Transportation Systems Conference (ITSC)}, 2019, pp. 419--424.

\bibitem{gangopadhyay2019identification}
B.~Gangopadhyay, S.~Khastgir, S.~Dey, P.~Dasgupta, G.~Montana, and P.~Jennings, ``Identification of test cases for automated driving systems using bayesian optimization,'' in \emph{2019 IEEE Intelligent Transportation Systems Conference (ITSC)}, 2019, pp. 1961--1967.

\bibitem{nabhan2019optimizing}
M.~Nabhan, M.~Schoenauer, Y.~Tourbier, and H.~Hage, ``Optimizing coverage of simulated driving scenarios for the autonomous vehicle,'' in \emph{2019 IEEE Int. Conf. Connect. Veh. Expo (ICCVE)}, 2019, pp. 1--5.

\bibitem{abbas2019safe}
H.~Abbas, M.~O’Kelly, A.~Rodionova, and R.~Mangharam, ``Safe at any speed: A simulation-based test harness for autonomous vehicles,'' in \emph{Cyber Physical Systems. Design, Modeling, and Evaluation. 7th International Workshop, CyPhy 2017, Revised Selected Papers}.\hskip 1em plus 0.5em minus 0.4em\relax Seoul, South Korea: Springer, 2019, pp. 94--106.

\bibitem{ding2020learning}
W.~Ding, B.~Chen, M.~Xu, and D.~Zhao, ``Learning to collide: An adaptive safety-critical scenarios generating method,'' in \emph{2020 IEEE Int. Conf. Intell. Rob. Syst. (IROS)}, 2020, pp. 2243--2250.

\bibitem{wei2024risk}
Z.~Wei, H.~Zhou, and R.~Zhou, ``Risk and complexity assessment of autonomous vehicle testing scenarios,'' \emph{Applied Sciences}, vol.~14, no.~21, p. 9866, 2024.

\bibitem{luo2025high}
X.~Luo, Z.~Wei, G.~Zhang, H.~Huang, and R.~Zhou, ``High-risk powered two-wheelers scenarios generation for autonomous vehicle testing using wgan,'' \emph{Traffic Injury Prevention}, vol.~26, no.~2, pp. 243--251, 2025.

\bibitem{shaw1998using}
P.~Shaw, ``Using constraint programming and local search methods to solve vehicle routing problems,'' in \emph{Principles and Practice of Constraint Programming—CP98}, ser. Lecture Notes in Computer Science, M.~Maher and J.-F. Puget, Eds.\hskip 1em plus 0.5em minus 0.4em\relax Berlin, Heidelberg: Springer, 1998, vol. 1520, pp. 417--431.

\bibitem{pisinger2010large}
D.~Pisinger and S.~Ropke, ``Large neighborhood search,'' in \emph{Handbook of Metaheuristics}, M.~Gendreau and J.-Y. Potvin, Eds.\hskip 1em plus 0.5em minus 0.4em\relax Boston, MA: Springer US, 2010, pp. 399--419.

\bibitem{ropke2006adaptive}
S.~Ropke and D.~Pisinger, ``An adaptive large neighborhood search heuristic for the pickup and delivery problem with time windows,'' \emph{Transp. Sci.}, 2006.

\bibitem{shi2023adaptive}
J.~Shi, H.~Mao, Z.~Zhou, and L.~Zheng, ``Adaptive large neighborhood search algorithm for the unmanned aerial vehicle routing problem with recharging,'' \emph{Appl. Soft. Comput.}, vol. 147, p. 110831, Nov. 2023.

\bibitem{kirkpatrick1983optimization}
S.~Kirkpatrick, C.~D. Gelatt, and M.~P. Vecchi, ``Optimization by simulated annealing,'' \emph{Science}, vol. 220, no. 4598, pp. 671--680, May 1983.

\bibitem{wang2022safety}
Y.~Wang, R.~Yu, S.~Qiu, J.~Sun, and H.~Farah, ``Safety performance boundary identification of highly automated vehicles: A surrogate model-based gradient descent searching approach,'' \emph{IEEE Transactions on Intelligent Transportation Systems}, vol.~23, no.~12, pp. 23\,809--23\,820, 2022.

\end{thebibliography}


% \end{IEEEbiography}


% % insert where needed to balance the two columns on the last page with
% %\newpage



%\vfill


\end{document}